\documentclass[10pt,conference]{IEEEtran}

\RequirePackage{snapshot}

\usepackage{caption}
\usepackage{subcaption}

\usepackage{moreverb}
\usepackage{verbatim}

\usepackage{acronym}
\usepackage{enumitem}
\usepackage{color}
\usepackage{authblk}

\usepackage{parskip}

\usepackage{latexsym}
\usepackage{tabularx}
\usepackage{cite} 
\usepackage[sharp]{easylist}
\usepackage{etoolbox}

\usepackage{siunitx}
\usepackage[usenames,dvipsnames,table]{xcolor}

\usepackage{amsmath}
\usepackage{amsfonts}
\usepackage{amssymb}
\usepackage{amsthm}

\usepackage{algorithmic}
\usepackage{algorithm}
\usepackage{cases}

\usepackage[colorinlistoftodos,disable]{todonotes}
\newcommand{\rc}[1]{{\color{black}#1}}

\usepackage{styles/mathdefs}

\renewcommand{\eqref}[1]{(\ref{eq:#1})}
\newcommand{\secref}[1]{\S\ref{sec:#1}}

\newcommand{\figref}[1]{Fig.~\ref{fig:#1}}
\newcommand{\tabref}[1]{Table~\ref{tab:#1}}

\let\OldEasylist\easylist
\let\OldEndEasylist\endeasylist
\renewenvironment{easylist}{%
    \OldEasylist%
    \ListProperties(Style**=\color{blue},Start1=1)%
}{%
    \OldEndEasylist%
}%

\begin{document}

\title{Driver Gaze Region Estimation\\Without Using Eye Movement}



\author{Lex Fridman}
\author{Philipp Langhans}
\author{Joonbum Lee}
\author{Bryan Reimer}
\affil{Massachusetts Institute of Technology (MIT)}

\maketitle

\begin{abstract}%

  Automated estimation of the allocation of a driver's visual attention may be a critical component of future Advanced
  Driver Assistance Systems. In theory, vision-based tracking of the eye can provide a good estimate of gaze
  location. In practice, eye tracking from video is challenging because of sunglasses, eyeglass reflections, lighting
  conditions, occlusions, motion blur, and other factors. Estimation of head pose, on the other hand, is robust to many
  of these effects, but cannot provide as fine-grained of a resolution in localizing the gaze. However, for the purpose
  of keeping the driver safe, it is sufficient to partition gaze into regions. In this effort, we propose a system that
  extracts facial features and classifies their spatial configuration into six regions in real-time. Our proposed method
  achieves an average accuracy of 91.4\% at an average decision rate of 11 Hz on a dataset of 50 drivers from an on-road
  study.
\end{abstract}

\begin{IEEEkeywords}%
Gaze classification, head pose estimation, driver distraction, driver assistance systems, on-road study.
\end{IEEEkeywords}

\section{Introduction}\label{sec:introduction}

Naturalistic driving studies have shown that a driver's allocation of visual attention away from the road is a critical
indicator of accident risk \cite{klauer2006impact}. Such work would suggest that a real-time estimation of driver's gaze
could be coupled with an alerting system to enhance safety when the driver is overly distracted or inattentive
\cite{coughlin2011monitoring}. High precision eye tracking that includes an estimate of pupil orientation in the vehicle
is costly and difficult. From an image processing perspective alone, difficulties involve the unpredictability of the
environment, presence of sunglasses occluding the eye, rapid changes in ambient lighting including situations of extreme
glare resulting from reflection, partial occlusion of the pupil due to squinting, vehicle vibration, image blur, poor
video resolution etc. For example, in \cite{yoshioka2014pupil}, a state-of-the-art algorithm for detecting pupils in the
presence of specular reflection achieves only an 83\% accuracy. In \cite{swirski2012robust}, an accuracy of 87\% is
achieved for a camera that is positioned off-axis, as it likely may need to be located inside a vehicle. Costs of high
resolution recording equipment and other computational requirements further enhance the difficulty of developing
practical, deployable solutions. Since pupil detection for eye tracking is often unreliable in real-world conditions,
the natural question we ask is: how well can we do without it? This is the question that motivated our efforts and
  makes this work distinct from a large body of literature on gaze estimation. We do not assume that the head pose
  vector is the same as the gaze vector (i.e., eye pose plus head pose). This assumption is especially invalid in the
  driving context because off-axis orientation of the eyes contribute significantly to a driver's gaze position. To answer this
  question, we draw upon 1,689,947 manually annotated images of drivers' faces. In this dataset, human annotators use
  eye and head orientation to label where the driver is looking. Our proposed system uses only positions of the head
  derived directly from facial video to predict the annotated labels. The large annotated dataset allows us to
  characterize how well a system is able to answer the following question: to what degree can the head pose vector be
  used to predict the gaze region under variable orientation of the eye? Put another way, this paper is a machine
  learning inquiry into the prediction of ocular movements and whether these movements can be linked to head pose in the
  design of a driver gaze classification system.

We propose a method for exploiting the correspondence between drivers' eye and head movement. These two
variables have been shown to be correlated but in complex ways that vary by operational mode (parked vs moving),
location of focus and other extrinsic and individual characteristics \cite{munoz2015analysis}. In terms of utilizing
head pose data as part of a gross distraction detection system, \cite{talamonti2013eye} showed that the farther off-axis
the focus point is (a conceptual overlap with reduced likelihood of adverse event detection), the more likely that a
glance will be accompanied by a head movement. We show that even small shifts in facial configuration is sufficiently
distinct for a classifier to accurately disambiguate head pose into one of six gaze regions.

\section{Related Work}\label{sec:related-work}

This work is related to three established areas of computer vision: facial feature extraction, head pose
estimation, and gaze tracking. This paper integrates cutting-edge algorithms and ideas borrowed and modified from each
of these fields in order to demonstrate effective eyes-free gaze classification in the wild (a large on-road driving
dataset).

The algorithm in \cite{kazemi2014one} uses an ensemble of regression trees for super-real-time face alignment. Our face
feature extraction algorithm drawn upon this method as it is built on a decade of progress on the face alignment problem
(see \cite{kazemi2014one} for a survey of this literature). The key contribution of the algorithm is an iterative
transform of the image to a normalized coordinate system based on the current estimate of the face shape. Also, to avoid
the non-convex problem of initially matching a model of the shape to the image data, the assumption is made that the
initial estimate of the shape can be found in a linear subspace.

Head pose estimation has a long history in computer vision. Murphy-Chutorian and Trivedi \cite{murphy2009head} describe
74 published and tested systems from the last two decades. Generally, each approach makes one of several assumptions
that limit the general applicability of the system in driver state detection. These assumptions include: (1) the video
is continuous, (2) initial pose of the subject is known, (3) there is a stereo vision system available, (4) the camera
has frontal view of the face, (5) the head can only rotate on one axis, (6) the system only has to work for one
person. While the development of a set of assumptions is often necessary for the classification of a large number of
possible poses, our approach skips the head pose estimation step (i.e., the computation of a vector in 3D space modeling
the orientation of the head) and goes straight from the detection of a facial features to a classification of gaze to
one of six glance regions. We believe that such a classification set is sufficient for the in-vehicle environment where
the overarching goal is to assess if the driver is distracted or inattentive to the driving context.

Video-based pupil detection and eye tracking approaches have been extensively studied. The main pattern recognition
approaches combine one or more features (corneal reflection, distinct pupil shape in combination with edge-detection,
characteristic light intensity of the pupil, and a 3D model of the eye) to derive an estimate of an individual's pupil,
iris, or eye position \cite{al2013eye}. In practice, for many of the reasons discussed earlier, eye tracking in the
vehicle context even for the experimental assessment of driver behavior is often inaccurate. Our approach focuses on the
head as the proxy for classifying broad regions of eye movement to provide a mechanism for real-time driver state
estimation while facilitating a more economical method of assessing driver behavior in experimental setting during
design assessment and safety validation.


\section{Dataset}\label{sec:dataset}

Training and evaluation is carried out on a dataset of 50 subjects drawn from a larger field driving study of 80
subjects that took place on a local interstate highway (see \cite{mehler2015multi} for detailed experimental methods). For each
subject, the collection of data was carried out in one of two vehicles: 2013 Chevrolet Equinox or Volvo XC60 (randomly
assigned). For the subset of 50 subjects considered: 26 drove the Chevrolet and 24 drove the Volvo. The drivers
performed a number of secondary tasks of varying difficulty including using the voice interface in the vehicle to enter addresses
into the navigation system and using the voice interface as well as manual controls to select phone numbers from a
stored phone list.

Both vehicles were instrumented with an array of sensors for assessing driver behavior. The sensor set included a camera
positioned on the dashboard of each vehicle that was intended to capture the driver's face for annotation of glance
behavior. The cameras were positioned off-axis to the driver and in slightly different locations in the two vehicles
(based upon features of the dashboard, etc.). As each driver positioned the seat (electronic in both vehicles)
differently the relative position of the driver in relation to the camera varied somewhat by subject and across each
driver over time (i.e., drivers move continuously in the seat, etc.). The camera was an Allied Vision Tech Guppy Pro
F-125 B/C, capturing grayscale images at a resolution of 800x600 and speed of 30fps. An initial analysis of the data was
conducted that included a double manual coding of driver glances transitions during secondary task periods (at a
resolution of sub-200ms) into one of 10 classes (road, center stack, instrument cluster, rearview mirror, left, right,
left blind spot, right blind spot, uncodable, and other). As detailed in \cite{mehler2015multi}, any discrepancies between
the two coders were meditated by an arbitrator.

\definecolor{lightHyo}{gray}{0.7}
\newcommand{\headerHyo}[1]{\rule[-1.2em]{0em}{3em}\renewcommand{\arraystretch}{1}\begin{tabular}[c]{@{}l@{}}#1\end{tabular}}
\renewcommand{\arraystretch}{1.5}
\begin{table}[h!]
  \centering
  \begin{tabular}{llll}
    \hline
    \headerHyo{Code Name} &
    \headerHyo{Total Frames\\Annotated} &
    \headerHyo{Frames With\\Face Detected} &
    \headerHyo{Detection\\Rate}\\
    \hline
    Road & 1,689,947 & 1,316,644 & 77.9\%\\\arrayrulecolor{lightHyo}\hline
    Instrument Cluster & 50,991 & 41,090 & 80.6\%\\\arrayrulecolor{lightHyo}\hline
    Left & 38,743 & 18,265 & 47.1\%\\\arrayrulecolor{lightHyo}\hline
    Rearview Mirror & 37,668 & 29,354 & 77.9\%\\\arrayrulecolor{lightHyo}\hline
    Center Stack & 28,339 & 24,835 & 87.6\%\\\arrayrulecolor{lightHyo}\hline
    Right & 15,073 & 11,071 & 73.4\%\\\arrayrulecolor{black}\hline
    Total & 1,860,761 & 1,441,259 & 77.5\%\\\arrayrulecolor{black}\hline
  \end{tabular}
  \caption{Dataset statistics for each class and in total. For each class, the table lists the total number of video
    frames and of those the number of frames where a single face was successfully detected.}
  \label{tab:dataset-stats}
\end{table}

In this paper, a broad random subset of data was drawn from the initial experiment and the ``left'' and ``left blind
spot'' classes / ``right'' and ``right blind spot'' classes were collapsed respectively in to ``left'' and
``right''. This merger was performed because the left/right blind spot regions (1) did not contain enough data and (2)
overlapped with the left/right regions respectively. \rc{Moreover, the selection of regions was made such that a human
  annotator can accurately label each gaze region by looking at the video frames. Such accurate robust annotation is
  central to our supervised learning approach since, by definition, standard classification requires non-overlapping
  classes.} Periods that were labeled ``uncodable'' and ``other'' were excluded. Subject pruning was completed to ensure
that every subject under consideration has sufficient training data for each of the six glance regions (road, center
stack, instrument cluster, rearview mirror, left, and right).

As shown in \tabref{dataset-stats}, the resulting dataset contains 1,860,761 images each annotated as belonging to one
of six glance regions. Approximately 90\% of those images belonged to the ``road'' class, with the fewest images (15,073)
belonging to the ``right'' class. The algorithm described in \secref{algorithm} is used for face detection. The gaze
region classification approach requires at least part of the face to be detected in the image. Therefore, in the
evaluation we include only the images where a face is detected. As the table shows, on average, a face is detected in
77.5\% of images.

\section{Features Extraction and Classification}\label{sec:algorithm}

\begin{figure*}
  \centering
  \includegraphics[width=\textwidth]{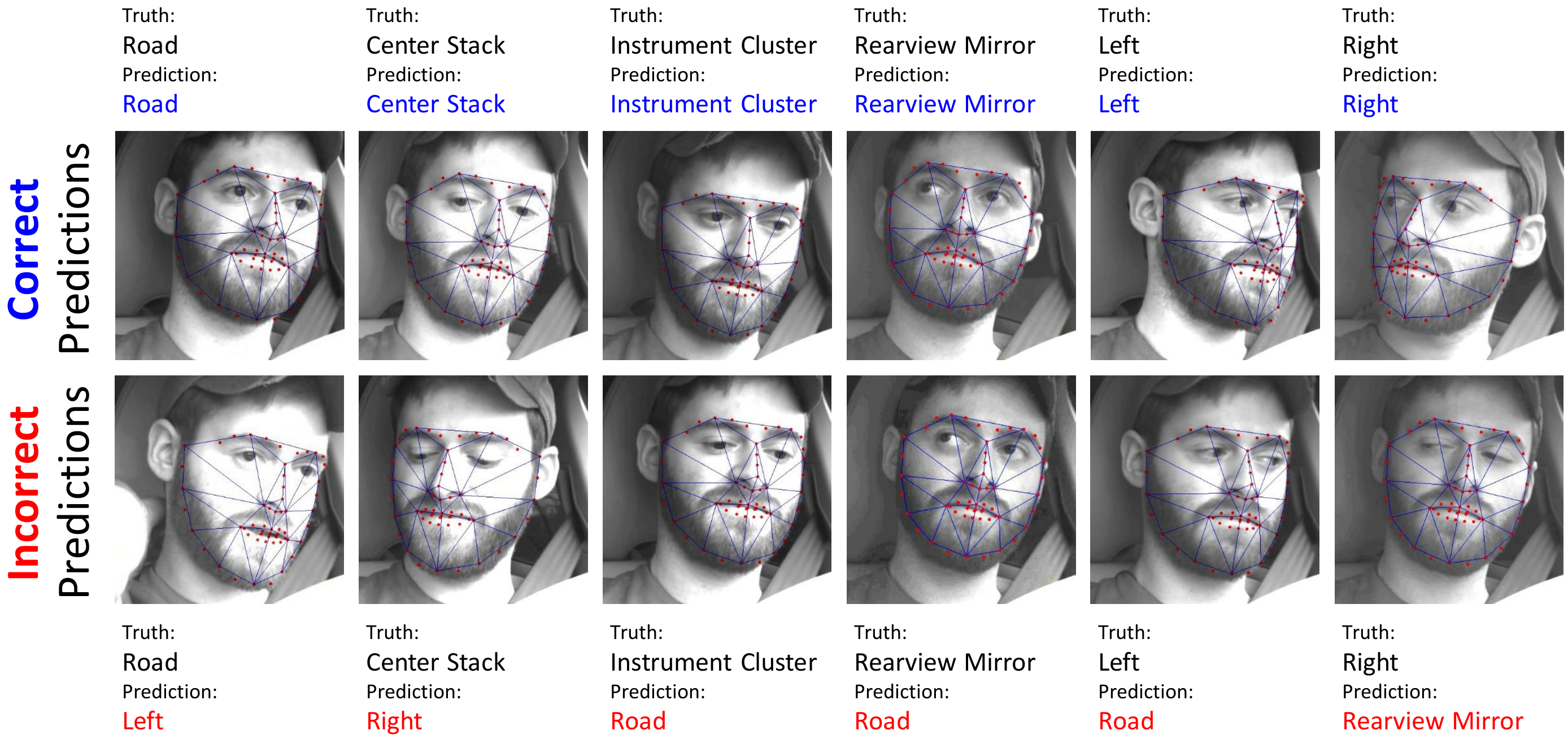}
  \caption{Representative examples for which the system predicted the glance region correctly (first row) and
    incorrectly (second row) for each of the six regions (six columns). Each image is labeled with 56 detected facial
    landmarks (red dots) and the Delaunay triangulation of a locally-optimal 19 landmark selection (blue lines). The
    images were light-corrected manually for presentation in this paper.}
  \label{fig:interesting-examples}
\end{figure*}

The steps in the gaze region classification pipeline are: (1) face detection, (2) face alignment, (3) feature
extraction, (4) feature normalization, (4) feature selection, (5) classification, (6) decision pruning. If the system
passes the first step (face detection) it will lead to a gaze region classification decision for every image fed into the
pipeline. In step 6, that decision may be dropped if it falls below a confidence threshold.

\subsection{Face Detection}

The face detector uses a Histogram of Oriented Gradients (HOG) combined with a linear SVM classifier, an image
pyramid, and sliding window detection scheme implemented in the DLIB C++ library \cite{dlib09}. The performance of this
detector has much lower false alarms rates than the widely-used default face detector available in OpenCV. For our
application, a false alarm is costly in both the case of a single face and multiple faces. In the former case, the error
ripples down to an almost certainly incorrect gaze region prediction. In the latter case, the video frame is dropped
from consideration, reducing the rate at which the system is able to make a decision.

\subsection{Estimation of Face Landmark Position}

Face alignment in our pipeline is performed on a 56-point subset from the 68-point Multi-PIE facial landmark mark-up
used in the iBUG 300-W dataset \cite{sagonas2013300}. These landmarks include parts of the nose, upper edge of the
eyebrows, outer and inner lips, jawline, and exclude all parts in and around the eye. The selected landmarks are shown
as red dots in \figref{interesting-examples}. The algorithm for aligning the 56-point shape to the image data uses a
cascade of regressors as described in \cite{kazemi2014one} and implemented in \cite{dlib09}. The two characteristics of
this algorithm most important to driver gaze localization is: (1) it is robust to partial occlusion and self-occlusion
and (2) its running-time is significantly faster than the 30 fps rate of incoming images.

\subsection{Feature Extraction, Normalization, and Selection}

The fact that a driver spends more than 90\% of their time looking at the road is used to normalize the spatial position
and orientation of facial landmarks such that they can be used to infer relative head movement across subjects. The
first 120 seconds (3,600 frames) of the video data for a subject are used to compute the average bounding box for the
face. All elements of the feature vector for a subject are normalized in relation to this bounding box. The first 112 elements in
the feature vector are the $(x, y)$ positions of the 56 facial landmarks. Of those 56 points, 19 are selected through
recursive feature elimination. Based on the face modeling approach in \cite{hassanpour2004delaunay}, a Delaunay
triangulation is computed over these 19 points and the three angles of each of the resulting triangles are added to the
feature vector. The triangulation is kept the same for all images, so for most of them the Delaunay property is not
maintained.

\subsection{Classification and Decision Pruning}\label{sec:classification}

Scikit-learn implementation of a random forest classifier \cite{scikit-learn} is used to generate a set of probabilities
for each class from a single feature vector. The probabilities are computed as the mean predicted class probabilities of
the trees in the forest. The class probability of a single tree is the fraction of samples of the same class in a
leaf. A random forest classifier of depth 30 with an ensemble of 1,000 trees is used for all experiments in
\secref{results}. The class with the highest probability is the one that the system assigns to the image as the
``decision''. The ratio of the highest probability to the second highest probability is termed the ``confidence'' of the
decision. A confidence of 1 is the minimum. There is no maximum. A confidence of ``infinity'' is assigned when all but
one of the classes have zero probability. The system only produces a decision when it is above a pre-specified
confidence threshold. The effect of this threshold is explored in \secref{confidence}.

\section{Results}\label{sec:results}

\subsection{Gaze Region Classes}

The evaluation of the gaze classification method is carried out over a 50 driver dataset described in
\secref{dataset}. One decision is made for every image where a face is detected without regard for temporal
information. We consider two ways to partition the driver gaze space. First, we consider the full six regions of (1)
road, (2) center stack, (3) instrument cluster, (4) rearview mirror, (5) left, and (6) right. Second, we combine regions
1, 3, 4, 5, and 6 together into a ``driving-related'' class, which results in a binary classification problem of
``driving-related'' versus center stack. The justification for this partitioning is that the ``driving-related'' regions
could be viewed as those not distracting to the driving task since they help the driver gain more information about the
driving environment. This is in line with the Alliance of Automobile Manufactures proposal \cite{driver2006statement}
for distraction evaluations based on glances to task-related areas such as displays (i.e., center stack). The left column
and right column of \figref{confusion} show the confusion matrices and accuracies for the six-region and the two-region
gaze classification problem, respectively. The classes in both cases are unbalanced since the ``road'' class accounts
for 90\% of the images. In order to evaluate the gaze classification system fairly, the size of the testing set for each
class is made equal.

\figref{interesting-examples} shows representative examples of where the system classifies gaze correctly and
incorrectly for each of the six gaze regions. The figure shows that the features used for classification do not capture
any information about eye or eyelid movement. Therefore, the system is robust to the driver wearing glasses or
sunglasses. The main takeaway from this figure is that the system correctly identifies gaze when the shift in attention
is accompanied with a movement of the head. It also highlights the fact that gaze region classification is a different
problem than head pose estimation because the movement of the head associated with a glance to a particular region is
small, but sufficiently distinct for the classifier to pick up, especially when given subject-specific training
examples.

\subsection{Global and User-Based Models}

\newcommand{\sixClassSize}{2.8in}
\newcommand{\twoClassSize}{2.7in}
\newcommand{\confusionVSpace}{\vspace{0.1in}}
\newcommand{\figConfusion}[2]{\includegraphics[width=#2]{images/single-experiments/#1/confusion/confusion1.pdf}}
\newcommand{\subfigConfusion}[6]{
  \begin{subfigure}{3.2in}
    \captionsetup{justification=centering}
    \figConfusion{#5}{#6}
    \caption{#1. Accuracy: #2\%}
    \label{fig:confusion-#4}
  \end{subfigure}
}
\newcommand{\confusionSpace}{\hspace{0in}}

\begin{figure*}[ht!]
  \centering
  \subfigConfusion{Global model, all predictions}{44.1}{44.1}{1}{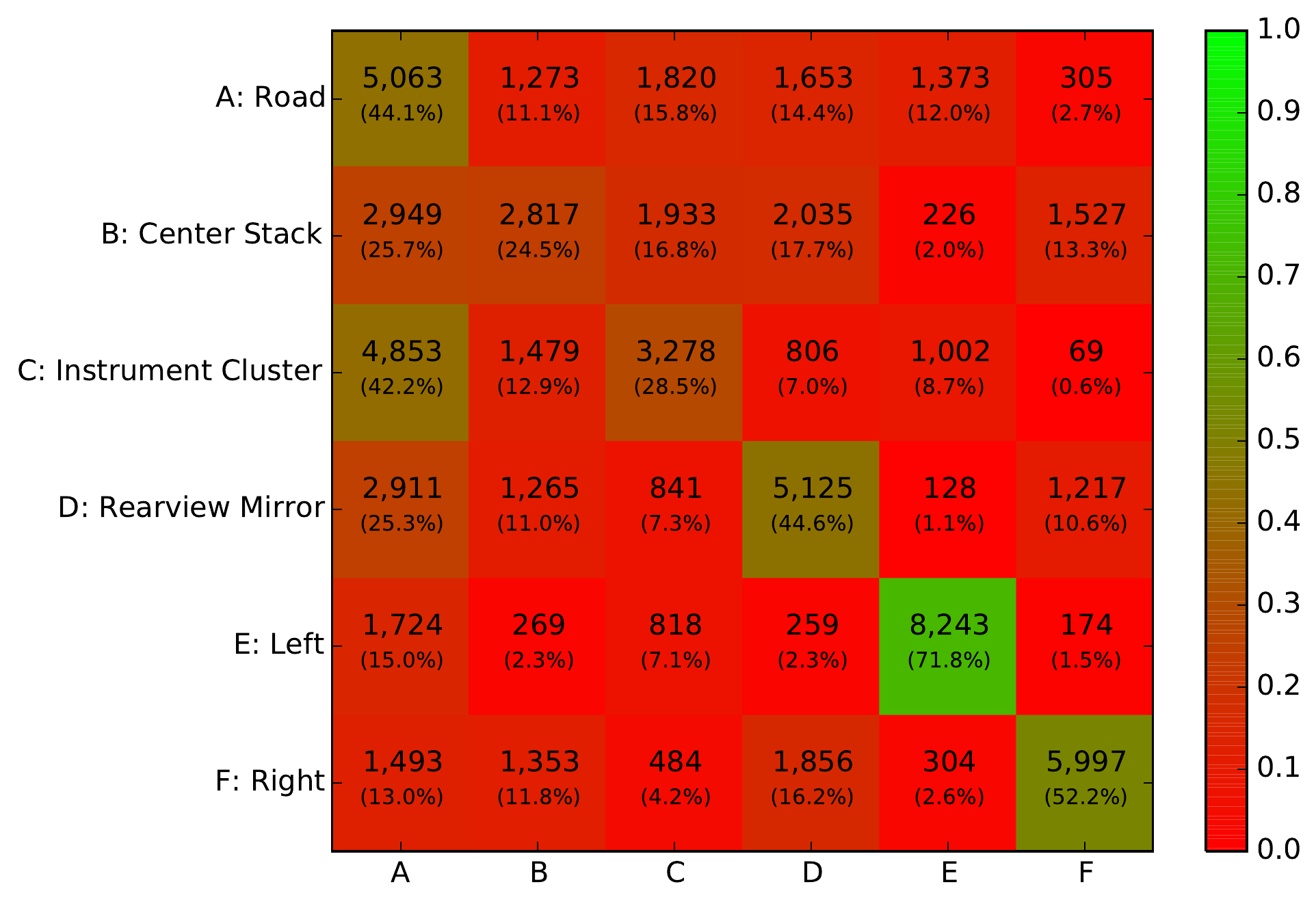}{\sixClassSize}\confusionSpace
  \subfigConfusion{Global model, all predictions}{61.8}{61.8}{2}{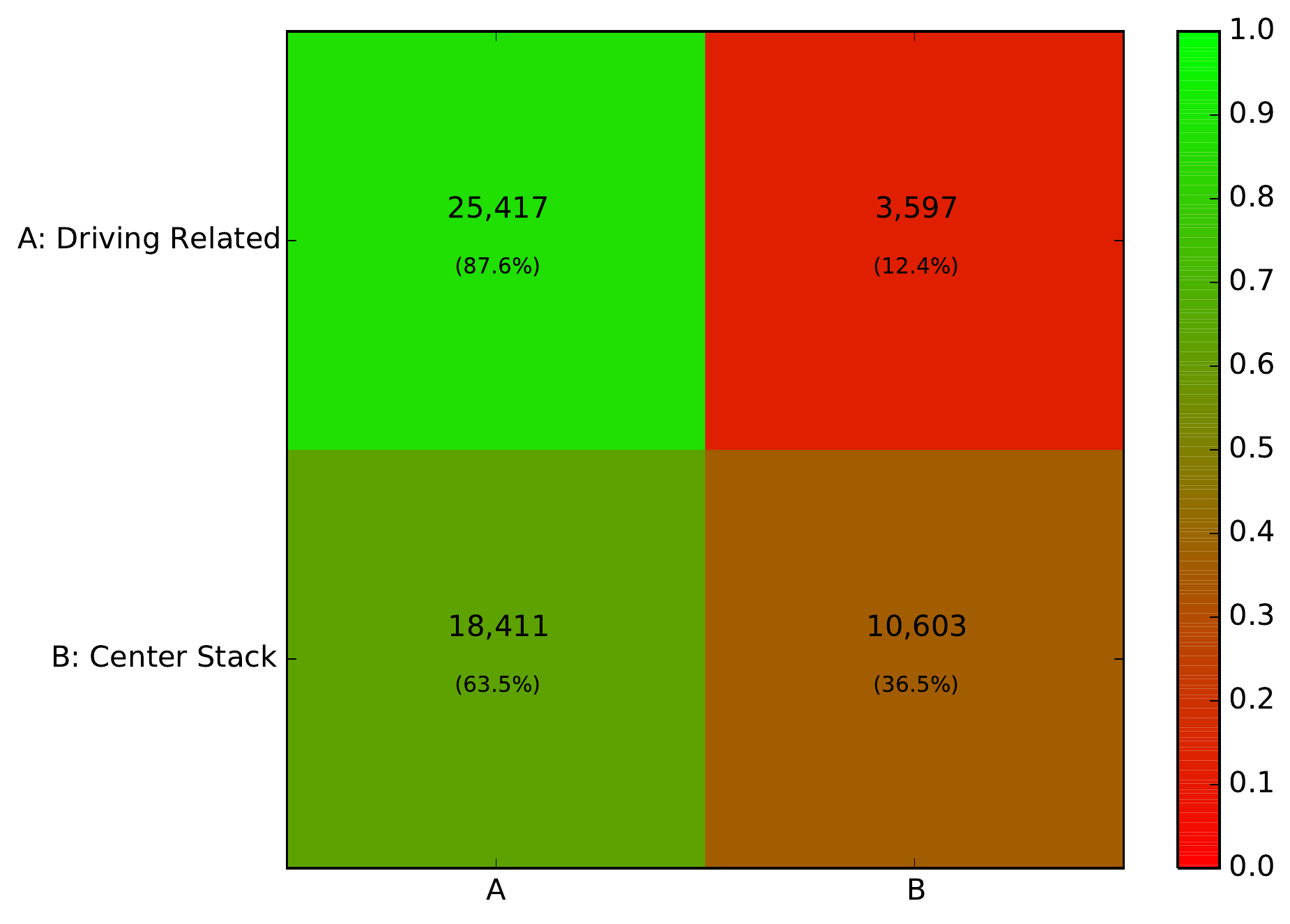}{\twoClassSize}\\\confusionVSpace
  \subfigConfusion{User-based model, all predictions}{65.0}{65.0}{5}{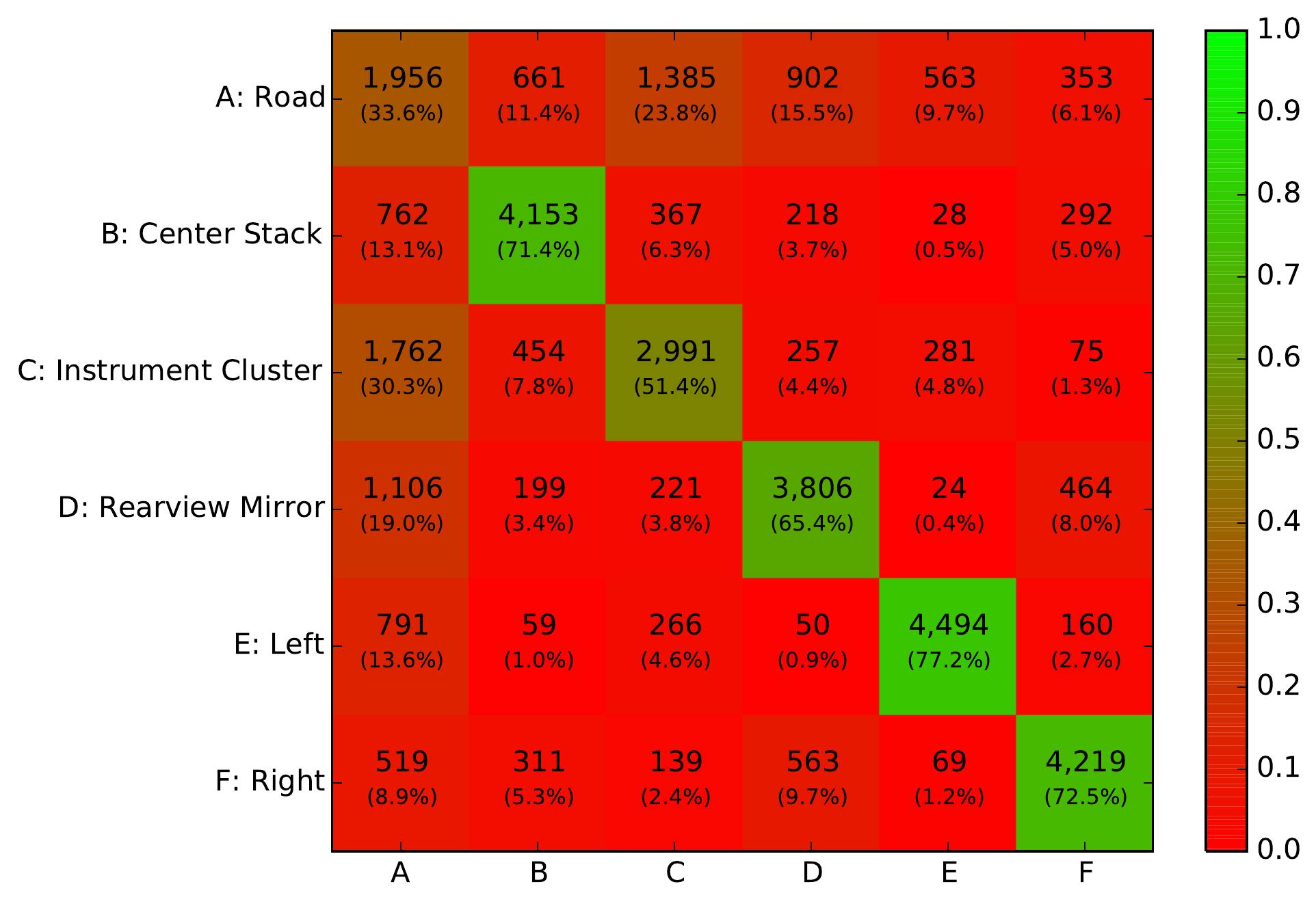}{\sixClassSize}\confusionSpace
  \subfigConfusion{User-based model, all predictions}{79.9}{79.9}{6}{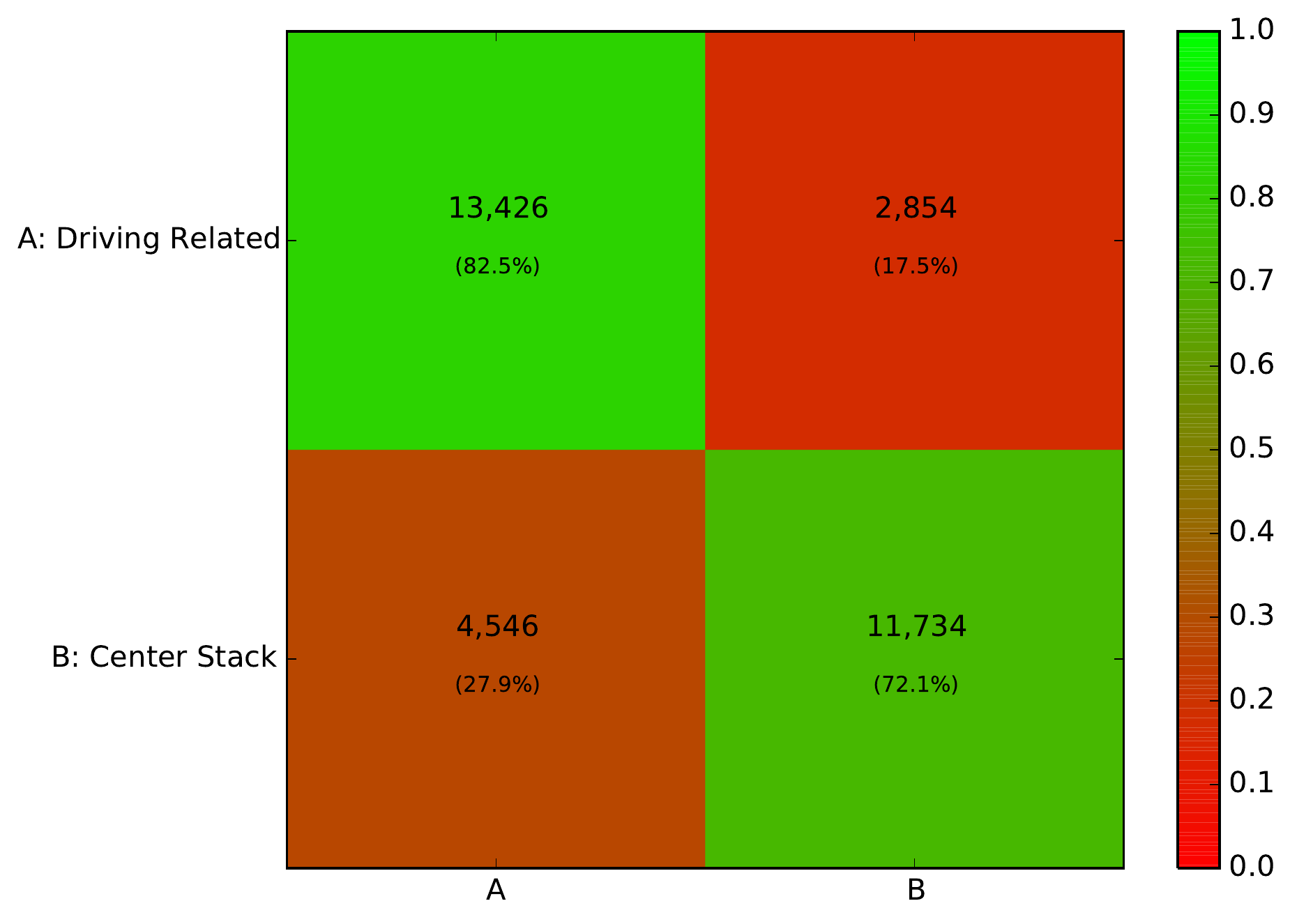}{\twoClassSize}\\\confusionVSpace
  \subfigConfusion{User-based model, confident predictions}{91.4}{86.8}{7}{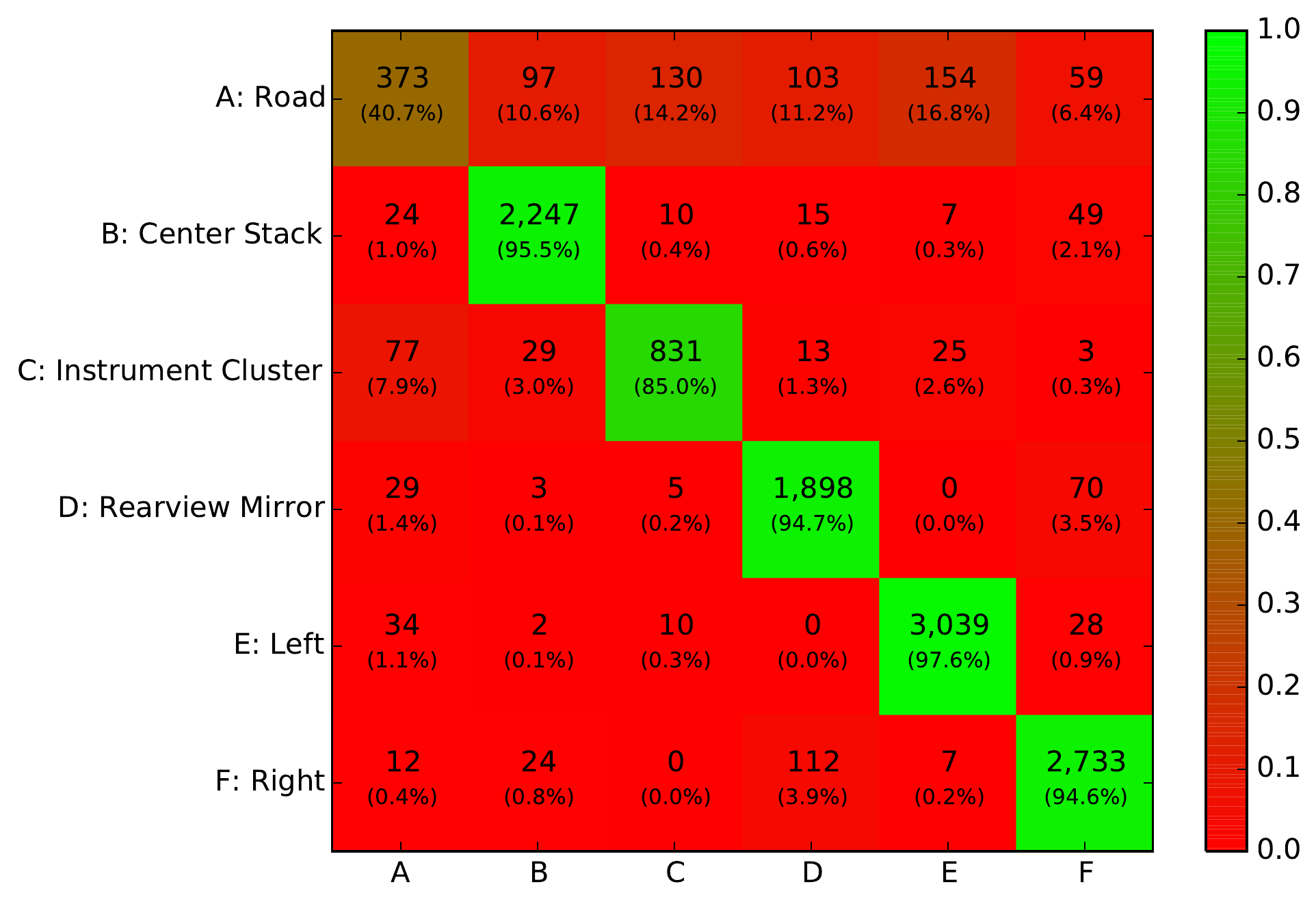}{\sixClassSize}\confusionSpace
  \subfigConfusion{User-based model, confident predictions}{92.5}{92.2}{8}{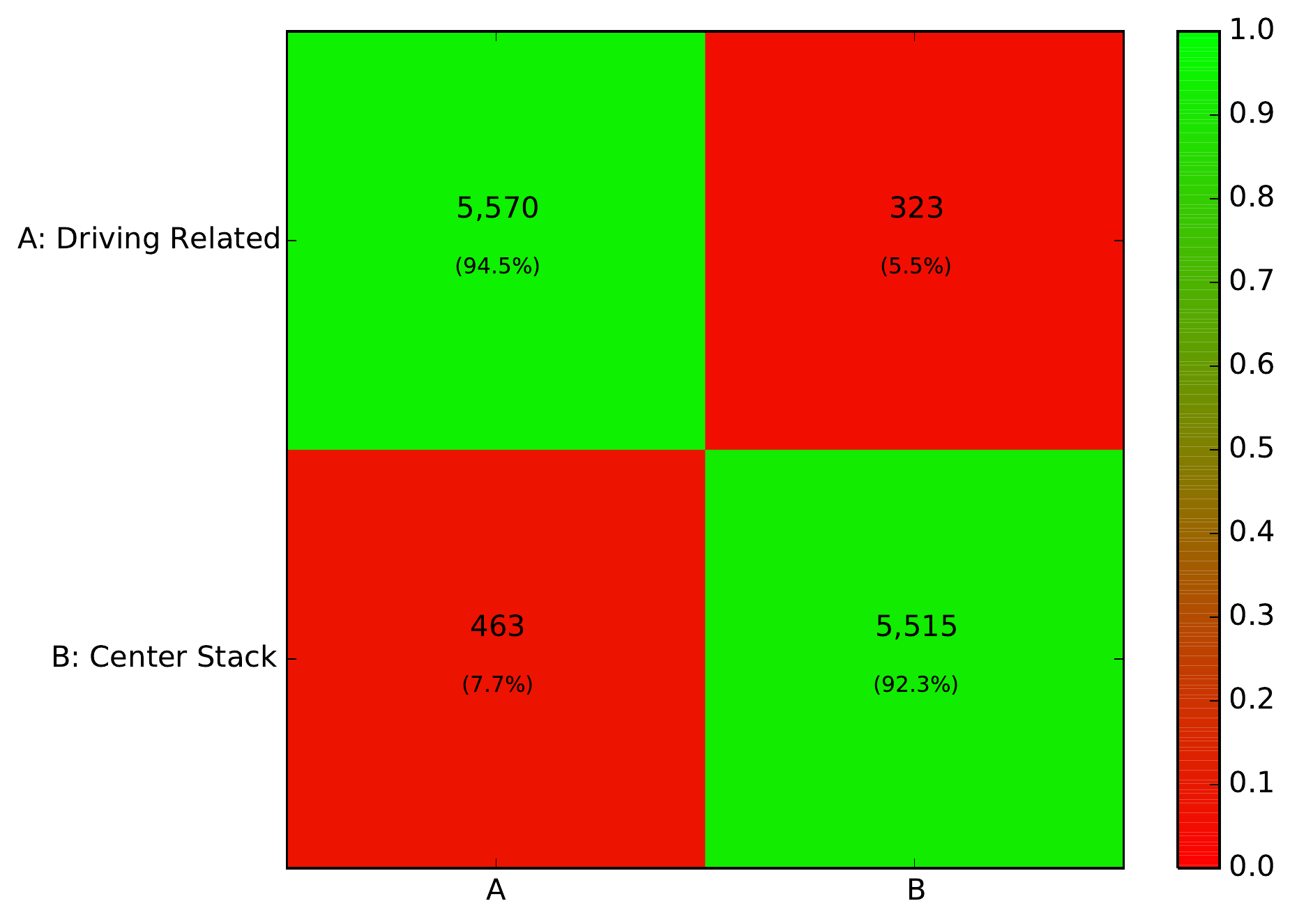}{\twoClassSize}\\
  \caption{Confusion matrices and accuracy for the six-region (left column) and two-region (right column) gaze
    classification problems. The first row shows the global model. The second and third rows show the user based model
    with and without the confidence-based decision pruning, respectively.}
  \label{fig:confusion}
\end{figure*}

\begin{figure}
  \centering
  \includegraphics[width=1.0\columnwidth]{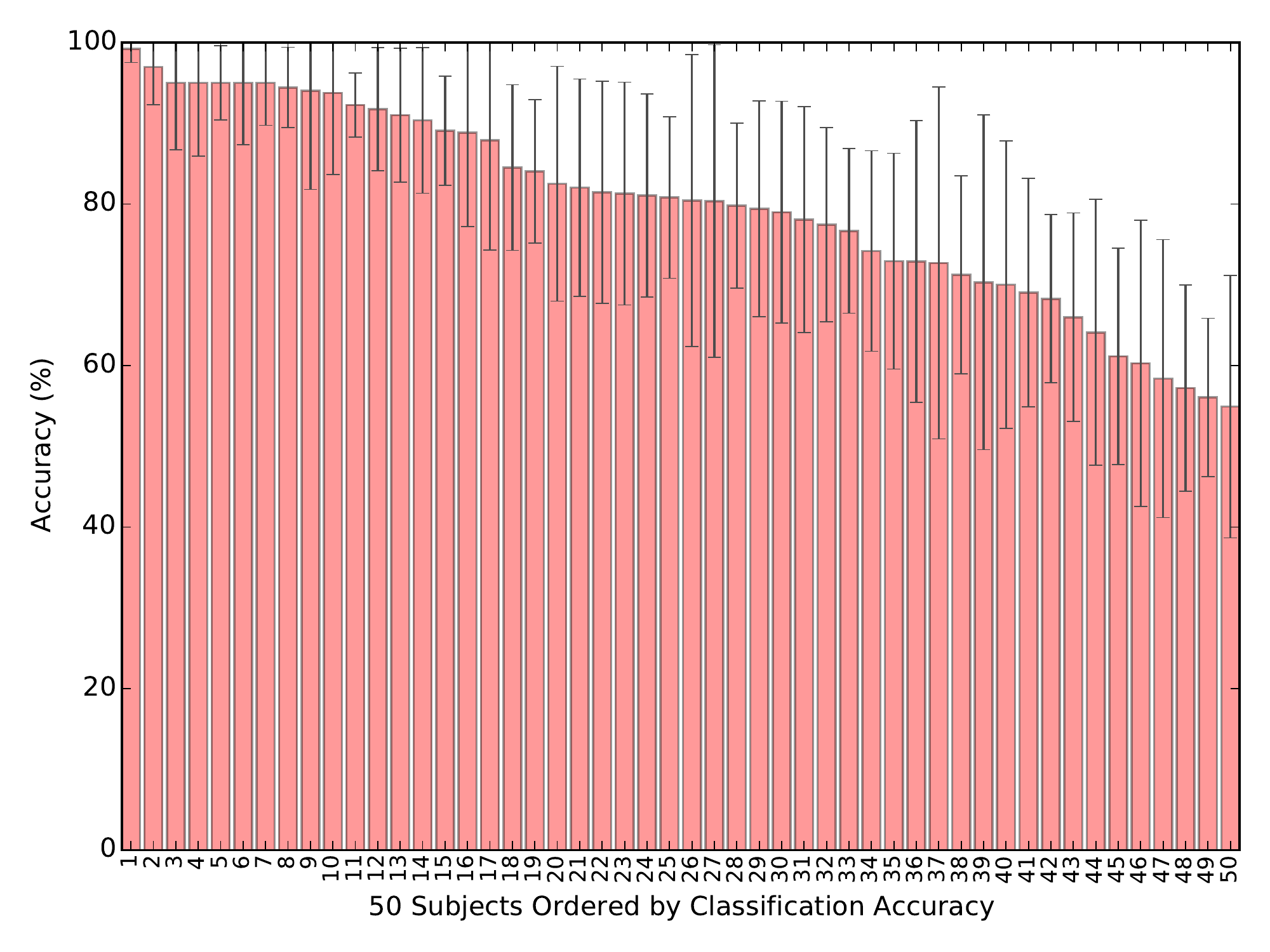}
  \caption{The variation in average classification accuracy for the two-class problem using the user-based model. The
    confusion matrix for this case is shown in \figref{confusion-6}. The errorbars show the standard deviation of
    accuracy for 100 random selections of training and testing sets for each of the 50 users.}
  \label{fig:per-user-accuracy}
\end{figure}

We consider two approaches for splitting the data into training and testing. The first method is an open world global
model that is trained on 40 subjects and tested on 10 subjects. The second method is a user-based model that is trained
for each subject on 90 consecutive images (corresponding to 3 seconds of video data) for each class. When
  the training and testing data is drawn from the same user, we ensure that the 90 images used for training come before the
  testing images and that there is at least a 30 second separation between the last training image and the first testing
  image.  The global model evaluates how well the proposed approach works without any training, calibration, or prior
knowledge of the user. A system based on this model could be placed inside the car and work almost right away without
any input needed from the driver. The user-based model requires an enrollment period which collects 3 seconds of video
data per class.

The first row of \figref{confusion} shows the performance of the system using the global model. An accuracy of 44.1\% is
achieved for the six class problem, and an accuracy of 61.8\% is achieved for the two class problem. Accuracy is
computed as the sum of correct decisions divided by the total number of decisions produced by the system over during the
evaluation. The testing set for both models is constructed to contain the same amount of continuous samples from each class. The
second row of \figref{confusion} shows the improved performance of the system using the user-based model. An accuracy of
65.0\% is achieved for the six class problem, and an accuracy of 79.9\% is achieved for the two class problem. The
evaluation of the global model and user based model is repeated 100 and 1,000 times over a random training-testing split
of the data to produce the mean and standard deviation seen in all the plots in this paper.

The performance of the user-based model is significantly better than the global model even though the amount of data
used for training in the former case is much smaller. This suggests that there is large variation between drivers in
terms of the relationship between their head movement and eye movement. This is confirmed by the plot in
\figref{per-user-accuracy} that shows the variation in classification accuracy of the two-class user-based model. The
confusion matrix for this case is shown in \figref{confusion-6}.

\subsection{Confidence-Based Decision Pruning}\label{sec:confidence}

\begin{figure}
  \centering
  \includegraphics[width=1.0\columnwidth]{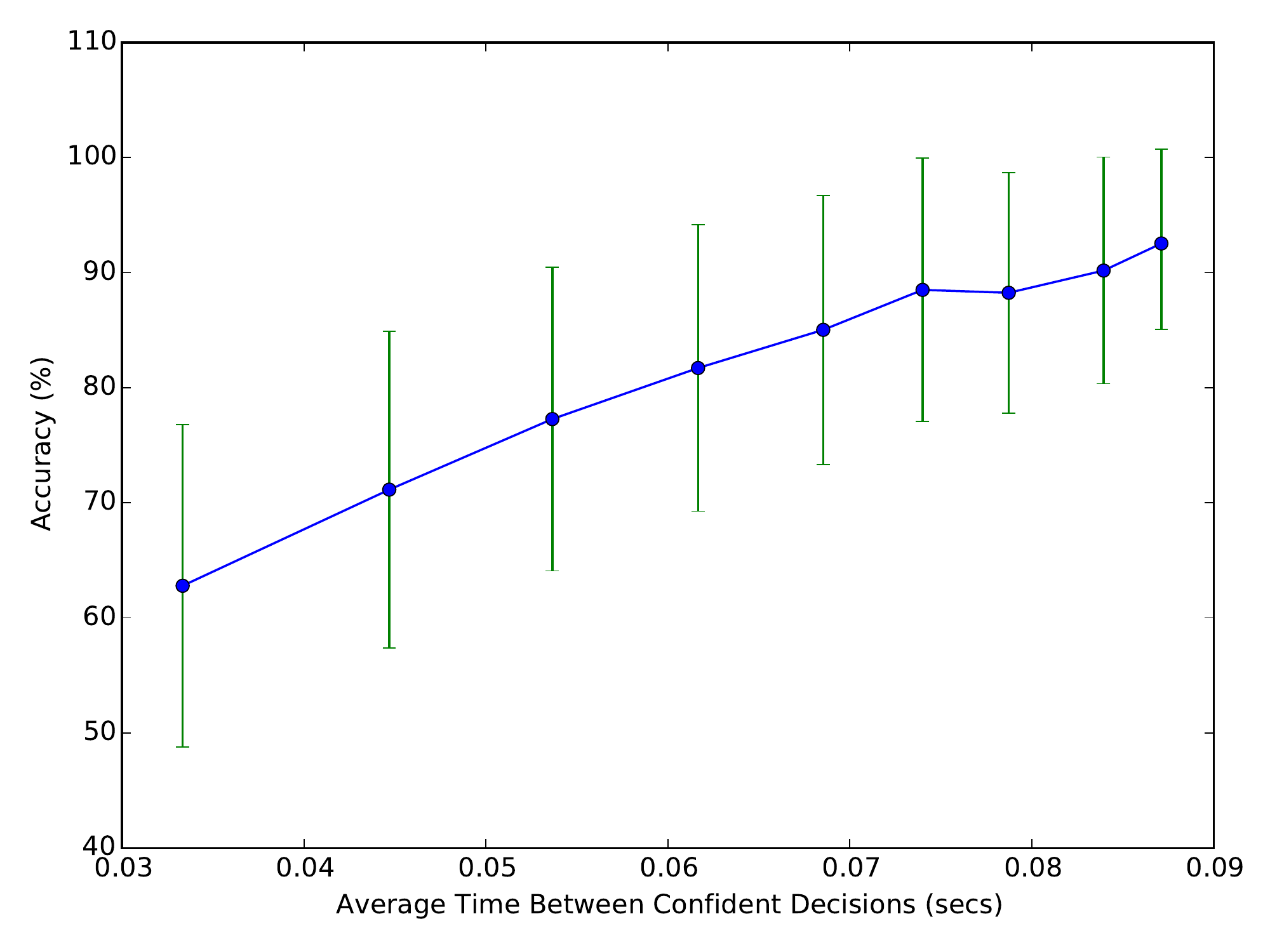}
  \caption{Six-class classification accuracy of the user-based model as the confidence threshold (and time between
    decisions) increases. The errorbars show the standard deviation of accuracy for 100 random selections of training
    and testing sets for each of the 50 subjects.}
  \label{fig:accuracy-vs-confidence}
\end{figure}

The main insight of this paper is that if we only use decisions that the system has high confidence in (see
\secref{classification}), the performance of the algorithm improves from 65.0\% to 91.4\% for the six class problem and
from 79.9\% to 92.5\% for the two-class problem as shown in the last row of \figref{confusion}. This is most likely due
to the nature of the relationship between head movement and eye movement. There appears to be a threshold in the spatial
configuration of facial features that delineates each of the six regions of driver attention from each other. Our
approach exploits this separability of confident decisions.

\figref{accuracy-vs-confidence} shows that as we increase the confidence threshold the accuracy of the system goes
up. The x-axis of this plot is the average time between confident decisions. As the confidence threshold increases, the
system does not produce classification decisions for some of the frames. However, the increase in the average decision
period is much slower than the increase in accuracy. Therefore, the cost of decreased decision rate is worth the big
increase in accuracy.

The steps in the gaze classification pipeline that have non-trivial running time are: face detection, face alignment,
Delaunay triangulation, and random forest classifier evaluation. Each of these steps runs under 10ms on a single core of
an Intel Core i5 2.4 GHz processor, suggesting the end-to-end gaze region classification pipeline can run in real-time
inside the car using only inexpensive consumer hardware.

\section{Conclusion}\label{sec:conclusion}

This paper shows that spatial configuration of facial landmarks provides sufficient discriminating information 
to accurately classify driver gaze into six gaze regions. The proposed system achieves an average accuracy of 91.4\% at
an average decision rate of 11 Hz for an on-road dataset of 50 subjects. Four observations are made about this
problem. First, building a subject-specific model (using 3 seconds of training data per class) improves classification
accuracy from 44.1\% to 65\%. Second, considering only confident classification decisions improves accuracy from 65\% to
91\%. Third, the problem of two region gaze classification (``driving-related'' versus center stack) that is
especially relevant to driver safety results in higher accuracy than the more general six-region classification
problem. Fourth, the classification accuracy varies significantly between subjects and within subjects. Our future work
will explore and exploit this inter-person and intra-person variation as it relates to the relationship between eye and
head movement.

\section*{Acknowledgment}

Support for this work was provided by the Santos Family Foundation, the New England University Transportation Center,
and the Toyota Class Action Settlement Safety Research and Education Program. The views and conclusions being expressed
are those of the authors, and have not been sponsored, approved, or endorsed by Toyota or plaintiffs’ class
counsel. Data was drawn from studies supported by the Insurance Institute for Highway Safety (IIHS).

\bibliographystyle{IEEEtran}
\bibliography{face,lex-fridman,agelab}

\end{document}